\documentclass[10pt,twocolumn,letterpaper]{article}

\usepackage{iccv}
\usepackage{times}
\usepackage{epsfig}
\usepackage{graphicx}
\usepackage{amsmath}
\usepackage{amssymb}
\usepackage{algorithm}
\usepackage{algorithmic}
\usepackage{multirow}
\usepackage{color}
\usepackage{color, colortbl}
\definecolor{Gray}{gray}{0.8}
\definecolor{Red}{rgb}{0.95,0.6,0.6}
\definecolor{LC}{rgb}{0.9,0.9,0.8}

\usepackage[pagebackref=true,breaklinks=true,letterpaper=true,colorlinks,bookmarks=false]{hyperref}

\iccvfinalcopy 

\def\iccvPaperID{523} 

\ificcvfinal\fi
\begin{document}

\title{Projection Bank: From High-dimensional Data to Medium-length Binary Codes}
\author{ Li Liu \qquad Mengyang Yu \qquad Ling Shao\\
Department of Computer Science and Digital Technologies\\
Northumbria University, Newcastle upon Tyne, NE1 8ST, UK\\
{\tt\small li2.liu@northumbria.ac.uk, m.y.yu@ieee.org, ling.shao@ieee.org}
}

\maketitle

\begin{abstract}
Recently, very high-dimensional feature representations, e.g., Fisher Vector, have achieved excellent performance for visual recognition and retrieval. However, these lengthy representations always cause extremely heavy computational and storage costs and even become unfeasible in some large-scale applications. A few existing techniques can transfer very high-dimensional data into binary codes, but they still require the reduced code length to be relatively long to maintain acceptable accuracies. To target a better balance between computational efficiency and accuracies, in this paper, we propose a novel embedding method called Binary Projection Bank (BPB), which can effectively reduce the very high-dimensional representations to \emph{medium-dimensional} binary codes without sacrificing accuracies. Instead of using conventional single linear or bilinear projections, the proposed method learns a bank of small projections via the max-margin constraint to optimally preserve the intrinsic data similarity. We have systematically evaluated the proposed method on three datasets: Flickr 1M, ILSVR2010 and UCF101, showing competitive retrieval and recognition accuracies compared with state-of-the-art approaches, but with a significantly smaller memory footprint and lower coding complexity.
\end{abstract}

\begin{figure}
  \centering
  \begin{tabular}{c}
    \includegraphics[width=0.45\textwidth]{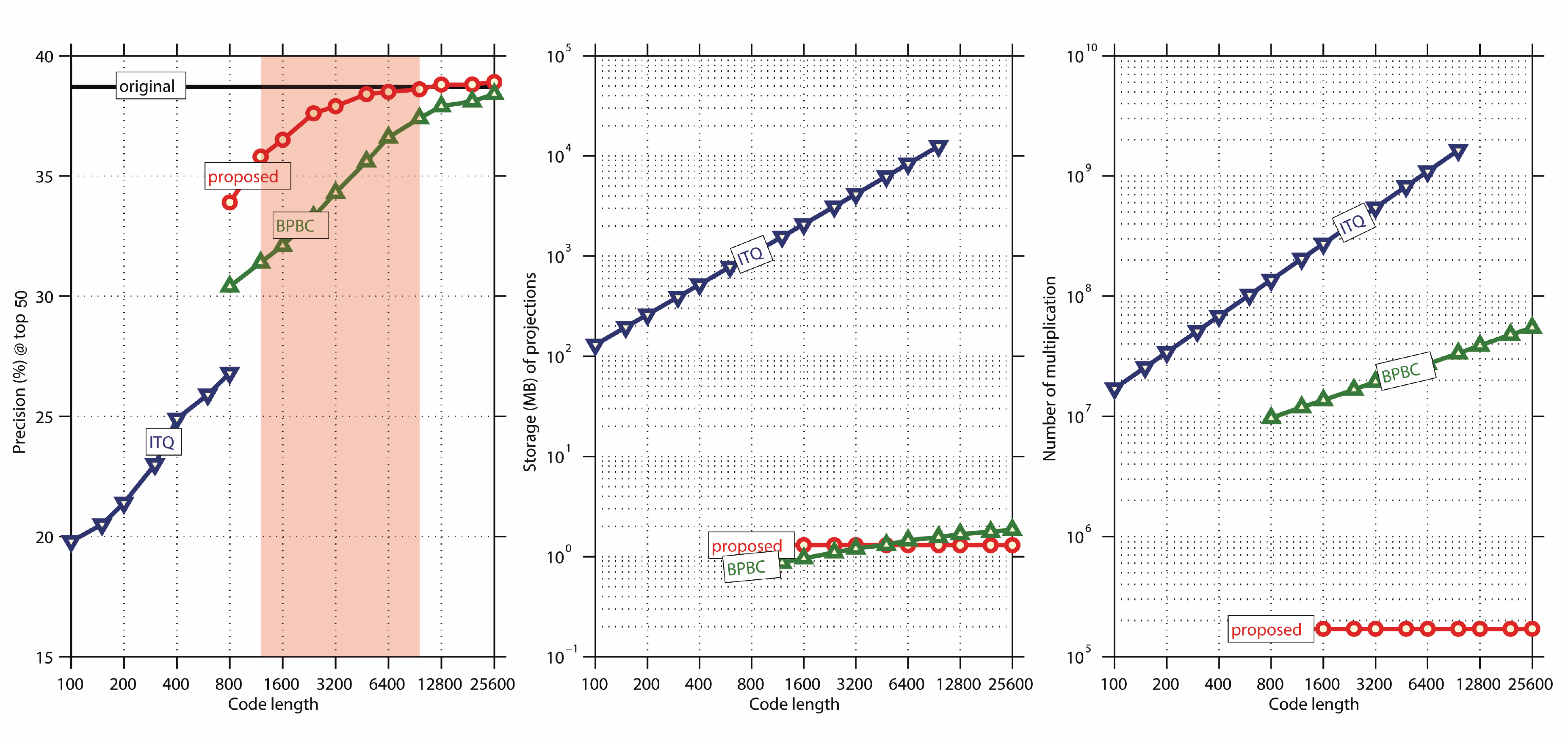} \\
    (a) \\
    \includegraphics[width=0.45\textwidth]{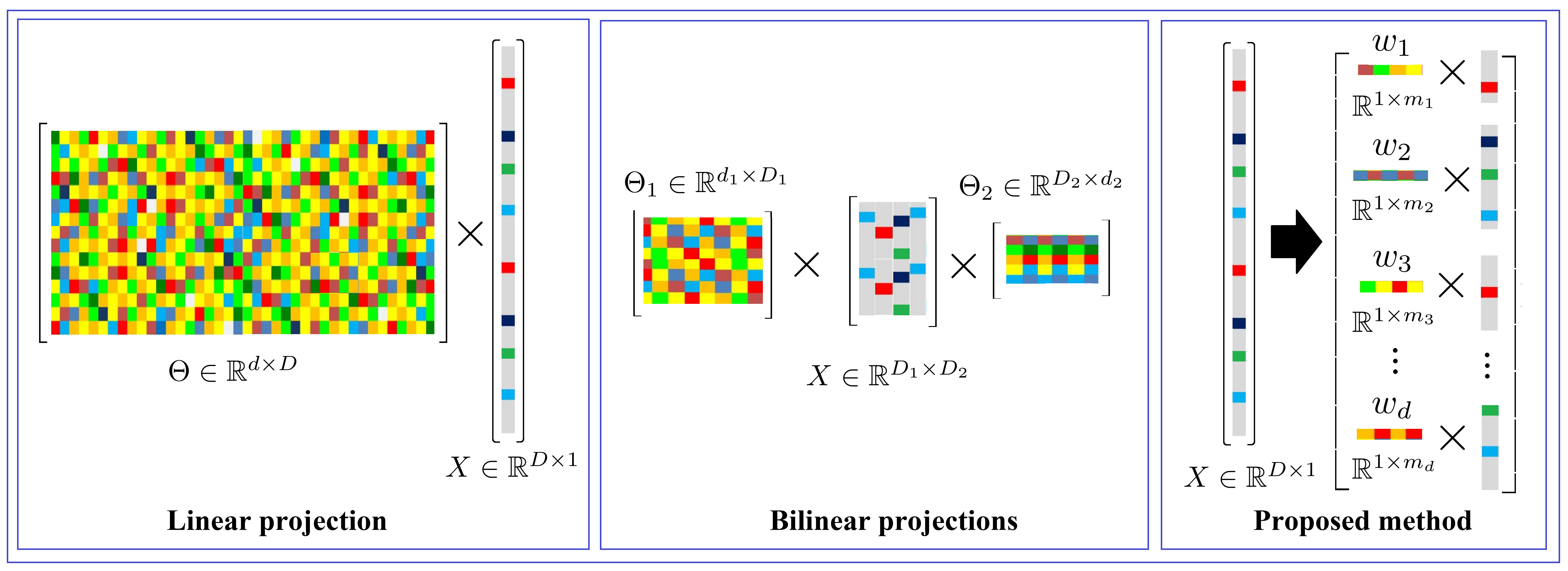} \\
    (b) \\
  \end{tabular}
  \caption{Comparison of the proposed method (projection bank) with state-of-the-art ITQ (linear projection) and BPBC (bilinear projections). (a-1) The comparison results for retrieval on the UCF 101~\cite{soomro2012ucf101} action dataset with around 10K videos. We use 1K videos as the query set and report the average semantic precisions at the top 50 retrieved points. Each video is represented via 170400-d FV (Original). Our goal is mainly to compare the results calculated on binary codes with \emph{medium-dimensions} (from 1000 bits to 10000 bits), where is shaded with red color in the figure. (a-2) The comparison of storage requirements (double precision) for three different projections. For ITQ, it is unfeasible to store the projections when code length exceeds 10000 bits. (a-3) The comparison of coding complexities of different projections. (b) Illustration of the three different coding methods.}
  \label{intro}
  \vspace{-2ex}
\end{figure}

\section{Introduction}
Recent research shows very high-dimensional feature representations, e.g., Fisher Vector (FV)~\cite{perronnin2010large,sanchez2011high,perronnin2007fisher} and VLAD~\cite{jegou2010aggregating},  can achieve state-of-the-art performance in many visual classification, retrieval and recognition tasks. Although these very high-dimensional representations lead to better results, with the emergence of massive-scale datasets, e.g., ImageNet \cite{deng2009imagenet} with around 15M images, the computational and storage costs of these long data have become very expensive and even unfeasible. For instance, if we represent 15M samples using 51200-dimensional FVs, the storage requirement of these data is approximately \nobreak5.6TB and it will need about $7.7\times10^{11}$ arithmetic operations measuring the Euclidean distance for image retrieval on these data. Considering the trade-off between computational efficiency and performance, it is desirable to embed the high-dimensional data into a reduced feature space. However, traditional dimensionality reduction methods such as PCA \cite{wold1987principal} are not suitable for large-scale/high-dimensional cases. The main reasons are: (1) Most dimensionality reduction methods are based on full-matrix linear projections, which need massive computational complexity and memory storage in high-dimensional reduction circumstances; (2) The reduced representations are usually real-valued vectors. When both dimensionality and the number of samples are large, real-valued codes severely limit the efficiency for retrieval and classification tasks compared with the binary codes. Thus, recent binarization approaches \cite{gong2011iterative,heo2012spherical,qinfast,salakhutdinov2009semantic,weiss2008spectral,liu2011hashing,lin2013compressed,charikar2002similarity,cai2011latent,yu2014circulant,perronnin2010large,sanchez2011high,gong2013learning,liu2015multiview} have been proposed to embed the original data into binary codes with a reduced dimension. The codes generated by these methods can be roughly divided into two groups, i.e., the short binary codes and the long binary codes.

\emph{Hashing short codes:} Most hashing-based approaches designed for fast searching always embed relatively low-dimensional representations like GIST~\cite{oliva2001modeling} into short binary codes (usually under 500 bits) without too much loss of information. However, recent sophisticated and state-of-the-art representations are always over ten thousand dimensions. Therefore, these hashing methods become not so effective and appropriate for the embedding of very high-dimensional data, since they cannot preserve the sufficient discriminative properties to maintain high performance if the length of the obtained binary codes is short according to~\cite{perronnin2010large,sanchez2011high}. Although some of hashing methods can theoretically generate long binary codes for high-dimensional data, the enormous computational load and memory usage make them unpractical. For instance, one of the state-of-the-art hashing methods, Iterative Quantization (ITQ)~\cite{gong2011iterative}, leads to unacceptable loss of retrieval accuracy compared with long codes over 1000 bits, meanwhile its computational cost becomes extremely high when the number of bits increases as shown in Fig.~\ref{intro}(a). \emph{Long binary codes:} Opposite to hashing short codes, a few methods~\cite{gong2013learning,yu2014circulant} have been specially introduced to convert the high-dimensional data to long binary codes. Among them, one representative binary coding method is Bilinear Projection-based Binary Codes (BPBC)~\cite{gong2013learning}, which can learn two rotate matrices for efficient binary coding. However, to minimize the loss of accuracies and achieve state-of-the-art performance, the left of Fig.~\ref{intro}(a) shows the length of binary codes generated by BPBC have to be long enough (i.e., over 10000 bits). Such long binary codes are still not fast enough for large-scale applications. Thus, how to learn \emph{medium-length binary codes} (i.e., between 1000 bits and 10000 bits) and still maintain the high accuracies becomes a challenging research topic.

Many existing binary code learning methods are based on a single linear projection matrix (e.g., random projection) to map the data from the high-dimensional space to a reduced space. No matter the training or the coding phase, the storage requirement for the single linear projection matrix remains a burden. Taking ITQ (PCA is involved as its first step) for an example, to reduce the 170400-dimensional FVs to 10000-dimension, the size of the given single projection matrix should be about 12.7GB and the number of multiplications for coding a new data sample is $1.7\times 10^{9}$.  Apparently, this kind of memory requirements and coding complexity of the linear projection is unrealistic for large-scale applications. To alleviate this weakness, a bilinear projection method~\cite{gong2013learning} has been proposed to effectively reduce the complexity of code learning compared with the linear one. The middle and right of Fig.~\ref{intro}(a) illustrate the memory usage and multiplications of coding respectively for the linear projection based ITQ and the bilinear projection based BPBC. Although the size of the projection matrix for the bilinear method is dramatically reduced, the coding complexity is still relatively high, especially when the dimensionality of original data goes high. Therefore, our target is to further reduce the coding complexity and produce medium-length codes without sacrificing the accuracy.

In this paper, we propose a novel binarization method for high-dimensional data. The proposed method first de-aggregates the original very high-dimensional representations into several groups of short representations according to their intrinsic data properties along the dimensions. After that, for each group of short representations, a small projection will be learned via the max-margin constraint to optimally preserve the data similarity. We denote our method as Binary Projection Bank (BPB), since a bank of small projections will be finally generated in our method instead of learning conventional linear or bilinear projections as illustrated in Fig.~\ref{intro}(b). The contributions of this paper include: (1) We propose a medium-length binary code learning method, which outperforms state-of-the-art linear and bilinear methods; (2) In spite of the reduced code length, our method only requires low and constant memory usage and coding complexity; (3) A kernelized version (KBPB) has also been proposed for better performance.

\section{Related Work}
There are a few works specifically focusing on high-dimensional data reduction. One of popular methods is Product Quantization (PQ)~\cite{jegou2011product}.  Prior to PQ, however, a random rotation is always needed to balance the variance of high-dimensional data according to~\cite{jegou2010aggregating}. As we discussed before, such rotation requires high computational complexity. Recently, an efficient high-dimensional reduction method based on feature merging~\cite{fulkerson2008localizing,jegou2008hamming,liu2011generalized}, termed Pseudo-supervised Kernel Alignment (PKA)~\cite{Liu2013merge}, has achieved good performance but with cheaper computation. Besides, aiming for large-scale tasks, some binary reduction techniques for high-dimensional data have also been introduced. Perronnin et al. \cite{perronnin2010large} proposed the ``$\alpha = 0$'' binarization scheme and compared with Locality Sensitive Hashing (LSH)~\cite{charikar2002similarity} and Spectral Hashing (SpH)~\cite{weiss2008spectral} on the compressed FVs. Hashing Kernel (HK)~\cite{shi2009hash} is utilized for high-dimensional signature compression as well in~\cite{sanchez2011high}. Most recently, the Bilinear Projection-based Binary Codes (BPBC) \cite{gong2013learning} is proposed to achieve more efficient binary coding. However, experiments in~\cite{perronnin2010large,sanchez2011high,gong2013learning} manifest these methods require very long codes to yield acceptable performance.

\section{Binary Projection Bank}
\subsection{Notation and Motivation}
We are given $N$ training data in a $D$-dimensional space: $\mathbf{x}_1, \cdots, \mathbf{x}_N \in \mathbb{R}^{D \times 1}$. The goal of this paper is to generate binary codes for each data point, for which the similarity and structure in the original data is preserved.

Traditional algorithms take $D$-dimensional data $\mathbf{x}$ as input and use a single projection $P \in \mathbb{R}^{D \times d}$ to form the linear prediction function $h(\mathbf{x}) = sgn(P^T \mathbf{x})$. Actually it can be regarded that projection matrix $P$ consists of $d$ linear classifiers (binary output: $0/1$) over the original feature space. However, for realistic high-dimensional data with noise and redundancy of dimension, learning single projections across the entire high-dimensional feature space is unwise and costs very high computational complexity. To tackle this problem, we aim to split the original high-dimensional feature space into $d$ subspaces by merging the similar dimensions together similar to \cite{Liu2013merge}. In this way, $d$ linear classifiers (i.e., projection vectors) will be explored since a subspace spanned by the dimensions with the similar property should only require one linear classifier. 
Hence, for each of subspace, only one small projection vector will be learned in our architecture. Table~\ref{comp} summarizes the resource requirements for different projection schemes.

\begin{table}
  \centering
\small
  \begin{tabular}{ccc}

   & Storage  & Coding complexity \\
   \hline
  Linear & $D  d$ & $D  d$ \\
  Bilinear & $D_1  d_1 + D_2  d_2$ & $D  (d_1 + d_2)$ \\
  Proposed & $D$ & $D$ \\
  \hline
\end{tabular}
  \caption{Storage and coding complexity of different projection schemes. The sizes of two matrices in the bilinear projection are $D_1 \times d_1$ and $D_2 \times d_2$. }\label{comp}
  \vspace{-3ex}
\end{table}

In this paper, we propose a Binary Projection Bank (\nobreak BPB) algorithm. To effectively decompose the original data space into subspaces, we first employ a K-means clustering scheme along dimensions. Particularly, we take $\mathbf{x}_1', \cdots, \mathbf{x}_D' \in \mathbb{R}^{1 \times N}$ as K-means input, which is the rows of matrix $[\mathbf{x}_1, \cdots, \mathbf{x}_N] \in \mathbb{R}^{D \times N}$, and divide them into $d$ clusters consisting of $\mathbf{x}_1'', \cdots, \mathbf{x}_D'' \in \mathbb{R}^{1 \times N}$ as illustrated:
\begin{equation}\label{}
\left[
\begin{array}{c}
 \mathbf{x}_1' \\
 \vdots \\
 \mathbf{x}_D' \\
\end{array}
\right]
  \underrightarrow{clustering} \left[\begin{array}{c}
                                             \left.
                                               \begin{array}{c}
                                                 \mathbf{x}''_1 \\
                                                 \vdots \\
                                                 \mathbf{x}''_{m_1} \\
                                               \end{array}
                                             \right\} \text{Cluster}~1 \\
                                             \vdots \\
                                             \left.
                                               \begin{array}{c}
                                                 \mathbf{x}''_{D-m_d+1} \\
                                                 \vdots \\
                                                 \mathbf{x}''_{D} \\
                                               \end{array}
                                             \right\} \text{Cluster}~d \\
                                           \end{array}
                                         \right].
\end{equation}
As mentioned in~\cite{Liu2013merge}, the decomposition via K-means can successfully preserve the intrinsic data structure and simultaneously group the dimension with the similar property together. We denote the number of the dimensions in the $p$-th cluster by $m_p$, where $p = 1, \cdots, d$. Then the subspace spanned by the dimensions in the $p$-th cluster is $\mathbb{R}^{m_p}$. In each subspace, a linear classifier is learned to generate one bit for the data. Hence, $D = m_1 + \cdots + m_d$ and $d$ is the reduced dimension. In the following, we give the detailed formulations and learning steps of BPB.


\subsection{Formulation of BPB}
To preserve the data similarity, we first construct a pseudo label $\ell_{ij}$ for each data pair $(\mathbf{x}_i, \mathbf{x}_j)$ according to their k-nearest neighbors in original data space as follows:
\begin{equation}\label{plabel}
  \ell_{ij} = \left\{\begin{array}{l}
                       +1,~  \text{if}~\mathbf{x}_i \in NN_k(\mathbf{x}_j) ~\text{or}~ \mathbf{x}_j \in NN_k(\mathbf{x}_i)  \\
                       -1,~  \text{otherwise}
                     \end{array}
  \right.,
\end{equation}
where $NN_k(\mathbf{x}_i)$ is the set of k-nearest neighbors of $\mathbf{x}_i$. Besides, we also define $\ell_{ii} = 1$ for $i=1, \cdots, N$.
In BPB learning phase, our goal is to minimize the distances of positive pairs and maximize the distances of the negative pairs in each subspace generated by K-means clustering.

For the $p$-th subspace, we use ${\mathbf{x}_{1(p)}, \cdots, \mathbf{x}_{N(p)}} \in \mathbb{R}^{m_p \times 1}$ to represent the data in this subspace. We tactfully transfer learning projections to learning linear classifiers via pair-wise labels generated from unlabeled data. By adopting linear classifier $f(\mathbf{x}) = \mathbf{w}^T \mathbf{x} - b$ similar to the SVM framework, positive pairs are positioned in the same side of the hyperplane while negative pairs are expected to be placed at different sides of the hyperplane.

In fact, we can denote $\widetilde{\mathbf{x}} = [\mathbf{x}^T, -1]^T$, then the classifier becomes $f(\widetilde{\mathbf{x}}) = [\mathbf{w}^T, b] \widetilde{\mathbf{x}}$. Therefore, it is equivalent to the linear classifier without the bias $b$.
In the following computation, we omit $b$ and the binary code for each data $\mathbf{x}_{(p)}$ in the $p$-th subspace can be acquired as follows:
\begin{equation}\label{}
h_p(\mathbf{x}_{(p)}) = sgn(\mathbf{w}_{(p)}^T \mathbf{x}_{(p)}),
\end{equation}
where $\mathbf{w}_{(p)}$ is the coefficient of the classifier for the $p$-th subspace. With the above requirement and the maximum margin criterion for the positive and negative pairs, we have the following optimization problem:
\begin{equation}\label{op1}
\begin{split}
&\min_{\mathbf{w}_{(p)}} \frac{1}{2} \|\mathbf{w}_{(p)}\|^2, \\
&\text{s.t.}~ \ell_{ij} \mathbf{w}_{(p)}^T \mathbf{x}_{i(p)} \cdot \mathbf{w}_{(p)}^T \mathbf{x}_{j(p)} > 1,~ i,j=1, \cdots, N,
\end{split}
\end{equation}
where $\frac{1}{2} \|\mathbf{w}_{(p)}\|^2$ is for the margin regularization. \emph{It is noticeable that if $i = j$, the constraint $\ell_{ij} \mathbf{w}_{(p)}^T \mathbf{x}_{i(p)} \cdot \mathbf{w}_{(p)}^T \mathbf{x}_{j(p)} > 1$ becomes $\ell_{ii} (\mathbf{w}_{(p)}^T \mathbf{x}_{i(p)})^2 > 1$, which strictly constrains that every point is out of the margin.} Using the hinge loss term, we can rewrite the optimization problem in (\ref{op1}) as the following objective function:
\begin{align}\label{loss}
  L(\mathbf{w}_{(p)}) = & \frac{1}{2} \|\mathbf{w}_{(p)}\|^2 \\
  \nonumber & + \lambda \sum_{i,j} \max(0, 1- \ell_{ij}  \mathbf{w}_{(p)}^T \mathbf{x}_{i(p)} \cdot \mathbf{w}_{(p)}^T \mathbf{x}_{j(p)}),
\end{align}
where $\lambda$ is the balance parameter to control the importance of the two terms. Since we cannot directly obtain the optimal $\mathbf{w}_{(p)}$ in our objective function,  a gradient descent scheme has to be applied here. Let us denote
\begin{equation}\label{hinge}
L_{ij}(\mathbf{w}_{(p)}) = \max(0, 1- \ell_{ij}  \mathbf{w}_{(p)}^T \mathbf{x}_{i(p)} \cdot \mathbf{w}_{(p)}^T \mathbf{x}_{j(p)}), \forall i,j.
\end{equation}
Taking the derivative of $L_{ij}(\mathbf{w}_{(p)})$ with respect to $\mathbf{w}_{(p)}$, we can easily obtain the gradient of $L_{ij}(\mathbf{w}_{(p)})$:
\begin{equation}\label{grad}
\nabla L_{ij}(\mathbf{w}_{(p)}) = \left\{
  \begin{array}{l}
    \mathbf{0},~ \text{if}~ 1- \ell_{ij}  \mathbf{w}_{(p)}^T \mathbf{x}_{i(p)} \cdot \mathbf{w}_{(p)}^T \mathbf{x}_{j(p)} \leq 0 \\
     - \ell_{ij} \left(\mathbf{x}_{i(p)} \mathbf{x}_{j(p)}^T + \mathbf{x}_{j(p)} \mathbf{x}_{i(p)}^T\right) \mathbf{w}_{(p)},~  \text{else}
  \end{array}
  \right..
\end{equation}
Note that in our implementation, if $1- \ell_{ij}  \mathbf{w}_{(p)}^T \mathbf{x}_{i(p)} \cdot \mathbf{w}_{(p)}^T \mathbf{x}_{j(p)} = 0$, we can set $\mathbf{w}_{(p)} \leftarrow \mathbf{w}_{(p)} + \Delta \mathbf{w}_{(p)}$, where $\Delta \mathbf{w}_{(p)}$ is a small nonzero random vector. The same scheme has also been used in \cite{kwak2008principal,wang2014fisher}.

Therefore, we utilize the gradient descent method and have the following update rule to optimize $\mathbf{w}_{(p)}$:
\begin{equation}\label{}
\begin{split}
\mathbf{w}_{(p)} &\leftarrow \mathbf{w}_{(p)} - \gamma \nabla L(\mathbf{w}_{(p)}) \\
&= \mathbf{w}_{(p)} - \gamma \Big(\mathbf{w}_{(p)} + \lambda \sum_{i,j} \nabla L_{ij}(\mathbf{w}_{(p)})\Big),
\end{split}
\end{equation}
where $\gamma$ is the step length.

We repeat the optimization problem in (\ref{op1}) for all the $d$ subspaces and concatenate the $d$ binary bits together to form the final binary code.
The final binary code for high-dimensional data $\mathbf{x}_i$ can be illustrated as:
\begin{equation*}\label{}
  [sgn(\mathbf{w}_{(1)}^T \mathbf{x}_{i(1)}), \cdots, sgn(\mathbf{w}_{(d)}^T \mathbf{x}_{i(d)})], ~i=1, \cdots, N.
\end{equation*}

\vspace{-2ex}\paragraph{Adaptive gradient descent (AGD):} Furthermore, for fast convergence, we also associate the optimization procedure with an adaptive step length. We first initialize $\gamma = 1$. For the $t$-th iteration, if $L(\mathbf{w}^{(t)}_{(p)}) \leq L(\mathbf{w}^{(t-1)}_{(p)})$, we enlarge the step length $\gamma \leftarrow 1.2 \gamma$ in the next iteration to accelerate the convergence, otherwise, we decrease $\gamma$ to its half size: $\gamma \leftarrow 0.5 \gamma$.
In the experiments, we also set an upper bound for the number of iteration for the gradient descent. Thus, we stop the iteration when the number of iteration reaches a maximum or the difference $|L(\mathbf{w}^{(t)}_{(p)}) - L(\mathbf{w}^{(t-1)}_{(p)})|$ is less than a small threshold.

\begin{figure}
  \centering
  \includegraphics[width=0.3\textwidth]{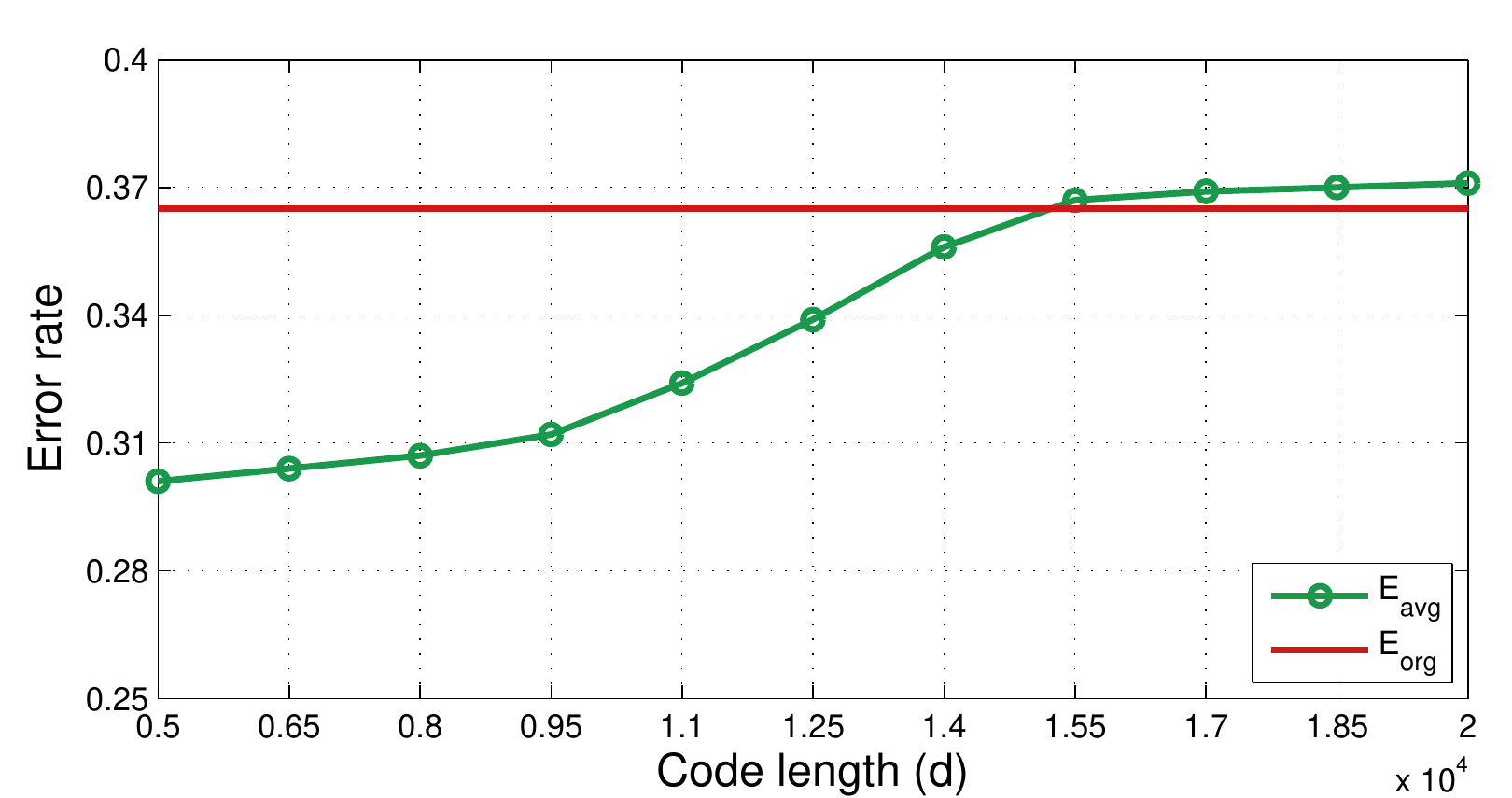}\\
  \caption{Comparison of the average pairwise error on subspaces and pairwise error on original data space with respect to reduced code length $d$ on the ILSVR2010 dataset with 64000-d FV. All the results are the means of 50 runs with 70-iteration of AGD.}\label{hingeloss}
  \vspace{-2ex}
\end{figure}


\vspace{-2ex}\paragraph{Average pairwise error of subspaces vs. pairwise error of uncompressed data:} We also analyze the empirical error rate of pairwise data in the projected space. Suppose $\mathbf{w}_{(p)}^*$ is the solution of minimizing $L(\mathbf{w}_{(p)})$ in Eq.(\ref{loss}) acquired by AGD for the $p$-th subspace, $p = 1, \cdots, d$. Then the average error rate on $N^{2}$ pair-wise data for all the subspaces is defined as
$E_{avg} =\frac{1}{d} \sum_{p=1}^d \frac{1}{N^2} \#\{(i,j)|sgn({\mathbf{w}_{(p)}^*}^T \mathbf{x}_{i(p)}) sgn({\mathbf{w}_{(p)}^*}^T \mathbf{x}_{j(p)}) \neq \ell_{ij}\}$, where $\#$ represents the cardinality of the set. On the other hand, we also compute the solution $\mathbf{w}^*$ of minimizing $L(\mathbf{w}) = \frac{1}{2} \|\mathbf{w}\|^2 + \lambda \sum_{i,j} \max(0, 1- \ell_{ij}  \mathbf{w}^T \mathbf{x}_{i} \cdot \mathbf{w}^T \mathbf{x}_{j})$, for the original $D$-dimensional data $\{\mathbf{x}_1, \cdots, \mathbf{x}_N\}$. Finally, we compare $E_{avg}$ with the error rate $E_{org} = \frac{1}{N^2}\#\{(i,j)|sgn({\mathbf{w}^*}^T \mathbf{x}_i) sgn({\mathbf{w}^*}^T \mathbf{x}_j) \neq \ell_{ij}\}$ in Fig.~\ref{hingeloss}. It is observed that $E_{avg}$ is lower than $E_{org}$ at the medium code length (under 10000 bits), which indicates that the data distribution in subspaces has much better separability than the one in the original space. However, when $d \rightarrow D$, the number of dimensions in each subspace will shrink to a very small value.  In this case, data in subspaces are difficult to be linearly separated by classifiers for the current \nobreak BPB. Thus, we will extend BPB to the version with non-linear kernels for better performance.

\subsection{Kernel BPB}

In this section, we introduce our algorithm with kernel functions, i.e., kernel BPB (KBPB), since the kernel method can theoretically and empirically be able to solve the linear inseparability problem mentioned in above. Although the kernel method would cost high computational complexity for high-dimensional data, in our method, the kernel function will only be performed in small subspaces which are spanned by the dimensions in each cluster. 

In the $p$-th subspace, suppose data are mapped to a Hilbert space by a mapping function $\phi$ and the kernel function $\kappa(\mathbf{x}_{i(p)}, \mathbf{x}_{j(p)}) = \phi(\mathbf{x}_{i(p)}) \cdot \phi(\mathbf{x}_{j(p)})$ is the inner product function in the Hilbert space. As defined in the Kernelized Locality-Sensitive Hashing (KLSH) \cite{kulis2012kernelized} and Kernel-Based Supervised Hashing (KSH) \cite{liu2012supervised}, we uniformly select $n$ samples from the training data (we call them \emph{basis samples}) to reduce the coding complexity of the kernel trick from $O(Nd)$ to $O(nd)$ (the effect on selection of $n$ will be discussed in experiments). Then we establish the prediction function $g_p$ with the kernel $\kappa$ as follows (without loss of generality, suppose the choices of basis samples are the first $n$ samples $\mathbf{x}_1, \cdots, \mathbf{x}_n$ for all the $d$ subspaces):
\begin{equation}\label{kobj}
  g_p(\mathbf{x}) = \sum_{i=1}^n a_i \kappa(\mathbf{x}_{i(p)}, \mathbf{x}) - b = \sum_{i=1}^n a_i \phi(\mathbf{x}_{i(p)})^T \phi(\mathbf{x}) -b,
\end{equation}
where $a_i \in \mathbb{R}$, $i=1, \cdots, n$ are the coefficients and $b \in \mathbb{R}$ is the bias. It is actually the linear classifier for the data $\phi(\mathbf{x})$ in the Hilbert space. The binary codes with sufficient information should be zero-centered \cite{liu2012supervised,kulis2012kernelized,gong2011iterative}, which renders that $\sum_{j=1}^n g_p(\mathbf{x}_{j(p)}) = 0$. To satisfy this condition, we set $b = \frac{1}{N} \sum_{j=1}^n \sum_{i=1}^n a_i \kappa(\mathbf{x}_{i(p)}, \mathbf{x}_{j(p)})$. Introducing $b$ in Eq. (\ref{kobj}), the prediction function becomes:
\begin{equation}\label{kobj}
  g_p(\mathbf{x})
   = \sum_{i=1}^n a_i \phi(\mathbf{x}_{i(p)})^T \phi(\mathbf{x}) - \sum_{i=1}^n a_i \mu_i
\end{equation}
where $\mu_i = \frac{1}{N} \sum_{j=1}^n\kappa(\mathbf{x}_{i(p)}, \mathbf{x}_{j(p)})$, $i= 1, \cdots, N$. It is easy to observe that $\sum_{i=1}^n a_i \phi(\mathbf{x}_{i(p)})$ is the coefficient vector of the hyperplane for the data $\phi(\mathbf{x})$ in the Hilbert space. With the similar constraints, we have the following optimization problem for KBPB:
\begin{align}\label{}
\min_{\mathbf{a}} & \frac{1}{2} \|\sum_{i=1}^n a_i \phi(\mathbf{x}_{i(p)})\|^2, \\
\nonumber\text{s.t.}~ & \ell_{ij} g_p (\mathbf{x}_{i(p)}) g_p (\mathbf{x}_{j(p)}) > 1, ~ i,j= 1, \cdots, n,
\end{align}
where $\mathbf{a} = (a_1, \cdots, a_n)^T$. Naturally, the corresponding objective function will be:
\begin{align*}
 \nonumber \widetilde{L}(\mathbf{a}) & = \frac{1}{2} \|\sum_{i=1}^n a_i \phi(\mathbf{x}_{i(p)})\|^2 \\
  \nonumber& \quad + \lambda \sum_{i,j} \max\left(0, 1- \ell_{ij} g_p (\mathbf{x}_{i(p)}) g_p (\mathbf{x}_{j(p)})\right) \\
  & = \frac{1}{2} \sum_{i=1}^n \sum_{j=1}^n a_i a_j \phi(\mathbf{x}_{i(p)})^T \phi(\mathbf{x}_{j(p)}) \\
 \nonumber & \quad + \lambda \sum_{i,j} \max\left(0, 1- \ell_{ij} g_p (\mathbf{x}_{i(p)}) g_p (\mathbf{x}_{j(p)})\right),
\end{align*}
where $\lambda$ is the balance parameter as in BPB.
Let us denote
$
  \widetilde{L}_1 (\mathbf{a}) = \frac{1}{2} \sum_{i=1}^n \sum_{j=1}^n a_i a_j \phi(\mathbf{x}_{i(p)})^T \phi(\mathbf{x}_{j(p)})
$
and
$
  \widetilde{L}_{ij} (\mathbf{a}) = \max\left(0, 1- \ell_{ij} g_p(\mathbf{x}_{i(p)}) g_p(\mathbf{x}_{j(p)})\right),~ i,j = 1, \cdots, n.
$
Then their derivatives with respect to $\mathbf{a}$ can be computed as:
\begin{equation*}
  \nabla \widetilde{L}_1 (\mathbf{a}) = \left[
                             \begin{array}{c}
                               a_1 \kappa(\mathbf{x}_{1(p)}, \mathbf{x}_{1(p)}) + \sum_{j \neq 1} a_j \kappa(\mathbf{x}_{1(p)}, \mathbf{x}_{j(p)}) \\
                               \vdots \\
                               a_n \kappa(\mathbf{x}_{n(p)}, \mathbf{x}_{n(p)}) + \sum_{j \neq n} a_j \kappa(\mathbf{x}_{n(p)}, \mathbf{x}_{j(p)}) \\
                             \end{array}
                           \right]
\end{equation*}
and
\begin{equation*}
\nabla \widetilde{L}_{ij}(\mathbf{a}) = \left\{
  \begin{array}{l}
    0,~ \text{if}~ 1- \ell_{ij}\left(\mathbf{a}^T \mathbf{k}_i - b\right)\left(\mathbf{a}^T \mathbf{k}_j - b\right) \leq 0 \\
    \\
     - \Big(\ell_{ij} (\mathbf{k}_i - \boldsymbol{\mu}) (\mathbf{k}_j - \boldsymbol{\mu})^T \mathbf{a} \\
      \quad + \ell_{ij}(\mathbf{k}_j - \boldsymbol{\mu}) (\mathbf{k}_i - \boldsymbol{\mu})^T \mathbf{a}\Big),~  \text{else}
  \end{array}
  \right.,
\end{equation*}
where $\boldsymbol{\mu}=[\mu_1, \cdots, \mu_n ]^T$ and $\mathbf{k}_i=[\kappa(\mathbf{x}_{1(p)}, \mathbf{x}_{i(p)}), \cdots, \kappa(\mathbf{x}_{n(p)}, \mathbf{x}_{i(p)}) ]^T$, $i=1,\cdots,n.$
Similar to BPB, if $1- \ell_{ij} g_p(\mathbf{x}_{i(p)}) g_p(\mathbf{x}_{j(p)}) = 0$, we can set $\mathbf{a} \leftarrow \mathbf{a} + \Delta\mathbf{a}$, where $\Delta\mathbf{a}$ is a small nonzero random vector.
Therefore, we have the update rule for KBPB as follows:
\begin{equation}\label{}
  \mathbf{a} \leftarrow \mathbf{a} - \gamma\Big(\nabla \widetilde{L}_1(\mathbf{a}) + \lambda \sum_{i,j} \nabla \widetilde{L}_{ij} (\mathbf{a})\Big),
\end{equation}
where $\gamma$ is the step length, which is also adaptively tuned by AGD.

Finally, having calculated the coefficients of the kernelized prediction function for all the $d$ subspaces, the binary codes for the original data $\mathbf{x}_i$ can be expressed as:
\begin{equation*}\label{}
  [sgn(g_1 (\mathbf{x}_{i(1)})), \cdots, sgn(g_d (\mathbf{x}_{i(d)}))], ~i=1, \cdots, N.
\end{equation*}

\section{Experiments}

\subsection{Large-scale image retrieval}
The proposed BPB and KBPB algorithms are first evaluated for the image similarity search task. Two realistic large-scale image datasets are used in our experiments: Flickr 1M and ILSVR2010. For \textbf{Flickr 1M}, we downloaded close to one million web images with 55 groups from Flickr inspired by \cite{wang2012learning,sanchez2011high}. For each image in Flickr 1M, we extract $128$-d SIFT  features in patches of $16\times16$ around interest points detected by \cite{lowe1999object}.  The \textbf{ILSVR2010}\footnote{http://www.imagenet.org/challenges/LSVRC/2010/index} dataset is a subset of the ImageNet \cite{deng2009imagenet} dataset and contains 1.2 million images from $1000$ categories. The publicly available dense $128$-d SIFT features \cite{deng2009imagenet} are used.

We represent each image in both datasets using two high-dimensional representations: Fisher Vector (FV) and VLAD. In respect to FV, the Gaussian Mixture Model is implemented on SIFT features with $250$ Gaussians for both datasets. In this way, the dimension of the final FV for each image is $2\times250\times128=64000$. While, for VLAD representations, the K-means clustering has been used to cluster the SIFT features into $250$ centers and aggregate them into VLAD vectors of $250\times128=32000$ dimensions. These VLAD vectors are also power and $\ell^{2}$ normalized \cite{perronnin2010improving}. In terms of both datasets, we randomly select $1000$ images as the query and the remaining images are regarded as the gallery database.  For evaluation, we first report the semantic precision at $50$ and $100$  retrieved images (according to the ground-truth) for both Flickr 1M and ILSVR2010, and then the precision-recall curves are illustrated as well. Additionally, we report the size of projection storage and the coding time (the average time used for each data) for some state-of-the-art methods. Our experiments are completed using Matlab 2014a on a server configured with a 6-core processor and 64GB of RAM running the Linux OS.

\vspace{-3ex}
\paragraph{Compared methods and settings:} In our experiments, we compare the proposed method with nine coding methods including four real-valued dimensionality reduction methods: Principal Component Analysis (PCA), the projection via Gaussian random rotation (RR), Product Quantization (PQ) \cite{jegou2011product} and Pseudo-supervised Kernel Alignment (PKA) \cite{Liu2013merge}, and five binary coding methods: the sign function binarization, ``$\alpha=0$'' binarization \cite{perronnin2010large}, Locality Sensitive Hashing (LSH) \cite{charikar2002similarity}, Spectral Hashing (SpH) \cite{weiss2008spectral}, Bilinear Projection-based Binary Codes (BPBC) \cite{gong2013learning} and Circulant Binary Embedding (CBE) \cite{yu2014circulant}. We use the publicly available codes of LSH, SpH, PQ , CBE and PCA, and implement RR, PKA and BPBC ourselves. Additionally, two natural baselines: randomly sampling the dimensions to form subspaces without replacement (RandST+BPB) and learning multiple bits with ITQ in each subspace (Kmeans+ITQ) are also included in our experiments. All of the above methods are then evaluated for compressing FV and VLAD representations into three different \emph{medium-lengthed codes}: $(8000, 6400, 4000; 4000, 3200, 2000)$. Considering the feasibility on the training phase of all the methods, in this experiment, 150K data are randomly selected from the gallery database of Flickr 1M and ILSVR2010 respectively to form the training set. Besides, we also randomly choose another 50K data samples from each of the datasets as a cross-validation set for parameter tuning. Under the same experimental setting, all the parameters used in the compared methods have been strictly chosen according to their original papers.

For the proposed BPB/KBPB, the pairwise label of each data pair is determined by their $100$ nearest neighbors. The balance parameter $\lambda$ for each dataset is selected from one of the values in the range of $[10^{-3},10^{2}]$, which yields the best performance on the cross-validation set. The maximum number of the iteration of AGD is fixed at $70$, which has been proved to converge well for the objective function. For KBPB, we adopt $n=1500$ as the number of basis samples for both Flickr 1M and ILSVR2010. We use the polynomial kernel $\kappa(\mathbf{x}_{i(p)}, \mathbf{x}_{j(p)})= (\mathbf{x}_{i(p)}^T\mathbf{x}_{j(p)}+1)^{\tau}$ and the RBF kernel $\kappa(\mathbf{x}_{i(p)}, \mathbf{x}_{j(p)})= \exp(-\|\mathbf{x}_{i(p)},\mathbf{x}_{j(p)}\|^{2}/\sigma^{2})$ to implement KBPB$^{1}$ and KBPB$^{2}$, respectively. The best value of $\tau$ for KBPB$^{1}$ is selected via cross-validation and the value of $\sigma$ for KBPB$^{2}$ is determined adaptively based on the method in~\cite{bai2010learning}. In fact, any kernel function satisfying the Mercer's condition can be used in KBPB. In BPB/KBPB, since the coding procedure in each subspace is independent, we implement the parallel computation scheme to speed up the training time. Considering the uncertainty of the K-means clustering, all the reported results by our methods are the averages of 50 runs.
\begin{table*}
\center
\newcommand{\tabincell}[2]{\begin{tabular}{@{}#1@{}}#2\end{tabular}}
\setlength{\baselineskip}{6pt}
\fontsize{6pt}{\baselineskip}\selectfont
\caption{Retrieval results (semantic precision) comparison on Flickr 1M with 64000-dimensional FV and 32000-dimensional VLAD.}
\label{table:t1}
\begin{tabular}{|c||c|c|c|c|c|c||c|c|c|c|c|c|}
\hline
\multirow{3}{*}{\textbf{Methods}} &
\multicolumn{6}{c||}{\textbf{Fisher Vector} (64000-d)} &
\multicolumn{6}{c|}{\textbf{VLAD} (32000-d)}\\

\cline{2-13}
 &
\multicolumn{3}{c|}{\textbf{Precision@top 50}} &
\multicolumn{3}{c||}{\textbf{Precision@top 100}}&
\multicolumn{3}{c|}{\textbf{Precision@top 50}} &
\multicolumn{3}{c|}{\textbf{Precision@top 100}}\\

\cline{2-13}
& \textbf{8000 bit} & \textbf{6400 bit} & \textbf{4000 bit} & \textbf{8000 bit} & \textbf{6400 bit} & \textbf{4000 bit} & \textbf{4000 bit} & \textbf{3200 bit} & \textbf{2000 bit}  & \textbf{4000 bit} & \textbf{3200 bit} & \textbf{2000 bit} \\
\hline
\hline
\rowcolor{LC}
Original &0.383&0.383 & 0.383&0.355  &0.355  &0.355 &0.370 & 0.370 & 0.370 & 0.339  & 0.339  & 0.339 \\
\hline
PCA  &0.371 &0.352 & 0.314 &0.354  &0.316 &0.284&0.365 &0.341 &0.302 &0.337 &0.295  &0.261 \\
PQ &0.160  &0.132 &0.121&0.154 &0.128 &0.111 &0.142 &0.114 &0.103 &0.138 &0.109  &0.090 \\
PKA &0.380 &0.374 &0.320 &0.352  &0.337 &0.301 &0.376 &0.366 &0.307 &  0.335&0.317  &0.276 \\
RR+PQ& 0.292 &0.279 &0.251 & 0.283 &0.268 &0.234 &0.291 &0.267 &0.233 &0.265  &0.249  &0.213 \\
\hline
Sign &0.281 &0.281 &0.281 &0.277  &0.277 &0.277 &0.271 &0.271 &0.271 &0.262 &0.262  &0.262 \\
$\alpha=0$&0.273 &0.273 &0.273&0.262 &0.262   &0.262  &- &- & - &-  &-  &- \\
LSH &0.319 &0.294 &0.270 &0.304  &0.289 &0.267 &0.315 &0.287 &0.250 &  0.287& 0.272 &0.244 \\
SpH &0.259 &0.288 &0.301  &0.244 &0.273 &0.296 &0.254 &0.267 &0.277 &0.225  &0.240  &0.265 \\
BPBC &0.343 &0.338 &0.314 &0.328  &0.303 &0.289 &0.328 &0.312 &0.294 &0.311  &0.281  &0.267 \\
CBE & 0.382 & 0.379 & 0.377 & 0.356  & 0.351 & 0.342 & 0.360 & 0.351 & 0.342 & 0.332  & 0.327  & 0.321 \\
\hline
\hline
Kmeans+ITQ(20bits)& 0.365 & 0.338 & 0.313 & 0.338  & 0.320 & 0.310 &  0.363 & 0.350 & 0.332 & 0.324 & 0.311  & 0.299 \\
 RandST+BPB & 0.352& 0.335 & 0.321 & 0.331  & 0.316 & 0.303 & 0.351 & 0.345 & 0.331 & 0.319 & 0.305  & 0.296 \\
\textbf{BPB}&  0.385 &0.381 &0.376 &0.356  &0.349 &0.345 &0.365 &0.357 &0.350 &0.338  &0.333  &0.329 \\
\textbf{KBPB}$^{1}$&   0.391 &0.388 &0.375 &0.358  &0.353 &0.347 &0.370 &0.362 &0.355 &0.342  &0.340  &0.332 \\
\rowcolor{Red}
\textbf{KBPB}$^{2}$ &0.398 &0.391 &0.379 &0.367  &0.359 &0.350 &0.378 &0.376 &0.363 &0.349  &0.345  &0.336 \\
\hline
\end{tabular}
\\ The ``Original'' indicates uncompressed FV/VLAD.  The ``Sign'' refers to directly using the sign function on original vectors. ``$\alpha=0$'' \cite{perronnin2010large} scheme is specifically designed for FV  and the dimension of reduced codes via ``$\alpha=0$'' is fixed at $(128+1)\times 250$=$32250$. KBPB$^{1}$ indicates KBPB with the polynomial kernel and KBPB$^{2}$ indicates KBPB with the RBF kernel. The results of BPB and KBPB are mean accuracies of 50 runs. For Original, PCA, PKA and RR, the Euclidean distance is used to measure the retrieval. For RR+PQ, the asymmetric distance (ASD)~\cite{jegou2011product} is adopted and Hamming distance is used for the rest of compared methods. Kmeans+ITQ(20bits) indicates using Kmeans to split the dimensions into subspaces and then apply ITQ to learning 20 bits codes for each subspace. RandST+BPB denotes randomly split the dimensions into subspaces without replacement and adopt BPB optimization scheme to learn codes.
\vspace{-1.5ex}
\end{table*}

\begin{table*}
\center
\newcommand{\tabincell}[2]{\begin{tabular}{@{}#1@{}}#2\end{tabular}}
\setlength{\baselineskip}{6pt}
\fontsize{6pt}{\baselineskip}\selectfont
\caption{Retrieval results comparison (semantic precision) on ILSVR2010 with 64000-dimensional FV and 32000-dimensional VLAD.}
\label{table:t2}
\begin{tabular}{|c||c|c|c|c|c|c||c|c|c|c|c|c|}
\hline
\multirow{3}{*}{\textbf{Methods}} &
\multicolumn{6}{c||}{\textbf{Fisher Vector} (64000-d)} &
\multicolumn{6}{c|}{\textbf{VLAD} (32000-d)}\\

\cline{2-13}
 &
\multicolumn{3}{c|}{\textbf{Precision@top 50}} &
\multicolumn{3}{c||}{\textbf{Precision@top 100}}&
\multicolumn{3}{c|}{\textbf{Precision@top 50}} &
\multicolumn{3}{c|}{\textbf{Precision@top 100}}\\

\cline{2-13}
& \textbf{8000 bit} & \textbf{6400 bit} & \textbf{4000 bit} & \textbf{8000 bit} & \textbf{6400 bit} & \textbf{4000 bit} & \textbf{4000 bit} & \textbf{3200 bit} & \textbf{2000 bit}  & \textbf{4000 bit} & \textbf{3200 bit} & \textbf{2000 bit} \\
\hline
\hline
\rowcolor{LC}
Original &0.214 &0.214 &0.214 & 0.175  &0.175 &0.175 &0.186 &0.186 &0.186 &0.149  &0.149  &0.149 \\
\hline
PCA &0.185 &0.177 &0.159 &0.154  &0.138 &0.125 &0.176 &0.160 &0.143 &  0.139& 0.116 &0.101 \\
PQ&0.157 &0.153 &0.138 & 0.132 &0.125 &0.110 &0.154 &0.142 &0.127 &0.115  &0.106  &0.089 \\
PKA &0.214 &0.205 &0.183 &0.179  &0.168 &0.155 &0.188 &0.172 &0.157 &  0.152&0.143  &0.124 \\
RR+PQ&0.182 &0.163 &0.157 &0.151  &0.137 &0.128 &0.174 &0.166 &0.151 &0.132  &0.117  &0.108 \\
\hline
Sign &0.152 &0.152 &0.152 &0.131  &0.131 &0.131 &0.141 &0.141 &0.141 & 0.109 &0.109  & 0.109\\
$\alpha=0$ &0.175 &0.175 &0.175 &0.144  &0.144 &0.144 &- &- & - &-  &-  &- \\
LSH &0.185 &0.171 &0.160 &0.162  &0.155 &0.143 &0.175 &0.159 &0.143 &  0.144&0.128  &0.117 \\
SpH &0.160 &0.169 &0.153 &0.139  &0.151 &0.123 &0.136 &0.152 &0.149 &  0.108&0.127  &0.120 \\
BPBC &0.187 &0.183 &0.176 &0.165  &0.153 &0.146 &0.180 &0.173 & 0.154&0.139  &0.127  &0.120 \\
 CBE & 0.214 & 0.207 & 0.197 & 0.178  & 0.164 & 0.155 & 0.184 & 0.174 & 0.167 &  0.145 & 0.138  & 0.131\\
\hline
\hline
 Kmeans+ITQ(20bits)&   0.201& 0.197 & 0.184 & 0.167  & 0.158 & 0.150& 0.172 & 0.165 & 0.155 & 0.137& 0.131  & 0.125 \\
 RandST+BPB&   0.202& 0.193 & 0.182 & 0.161 & 0.149 & 0.140 & 0.175 & 0.164 & 0.155 & 0.138 & 0.129  & 0.121 \\
\textbf{BPB} &0.210 &0.205 &0.197 &0.175  &0.166 &0.158 &0.186 &0.177 &0.165 &0.148  &0.142  &0.134\\
\textbf{KBPB}$^{1}$&0.218 &0.208 &0.201 &0.177  &0.172 &0.163 &0.190 &0.181 &0.171 &0.151  &0.145  &0.137 \\
\rowcolor{Red}
\textbf{KBPB}$^{2}$&0.226 &0.221 &0.208 &0.181  &0.176 & 0.168&0.198 &0.190 &0.180 &0.157  &0.150  &0.143 \\
\hline
\end{tabular}
\vspace{-3.5ex}
\end{table*}

\vspace{-3ex}
\paragraph{Results comparison:} We list the retrieval results comparison of different methods at top 50 and 100 retrieval results on the Flickr 1M and ILSVR2010 datasets in Table~\ref{table:t1} and Table~\ref{table:t2}, respectively. Generally, FV gains slightly better results than VLAD on both datasets. Meanwhile, the accuracies on the ILSVR2010 dataset are lower than those on the Flickr 1M dataset, since there are more categories and larger intra-class variations in ILSVR2010. It is noticeable that PQ achieves the low precision on Flickr 1M, while RR+PQ can lead to more reasonable results. The reason is that for high-dimensional representations, there may exist unbalanced variance that influences the performance. Thus, randomly rotating the high-dimensional data prior to PQ\footnote{In \cite{jegou2011product}, PQ can achieve competitive results without random rotation. However, they focus on relatively low-dimensional SIFT/GIST features whose variance already tends to be roughly balanced.} is recommended in \cite{jegou2010aggregating}. Nevertheless, due to that the images in ILSVR2010 are textured with the dominant object which leads to relatively balanced variance, the basic PQ can achieve modest results on ILSVR2010. PCA and PKA have remarkable accuracies as real-valued compression techniques on both datasets and  CBE is regarded as the strongest baseline of binary coding methods according to its performance. LSH, SpH and the ``$\alpha=0$'' scheme can obtain similar results on both datasets and using sign function directly on uncompressed FV/VLAD is proved to be the worst binarization method. Additionally, Kmeans+ITQ(20bits) can achieve slightly better performance than RandST+BPB, but both significantly lower than BPB.

\begin{figure}
  \centering
  \includegraphics[width=0.4\textwidth]{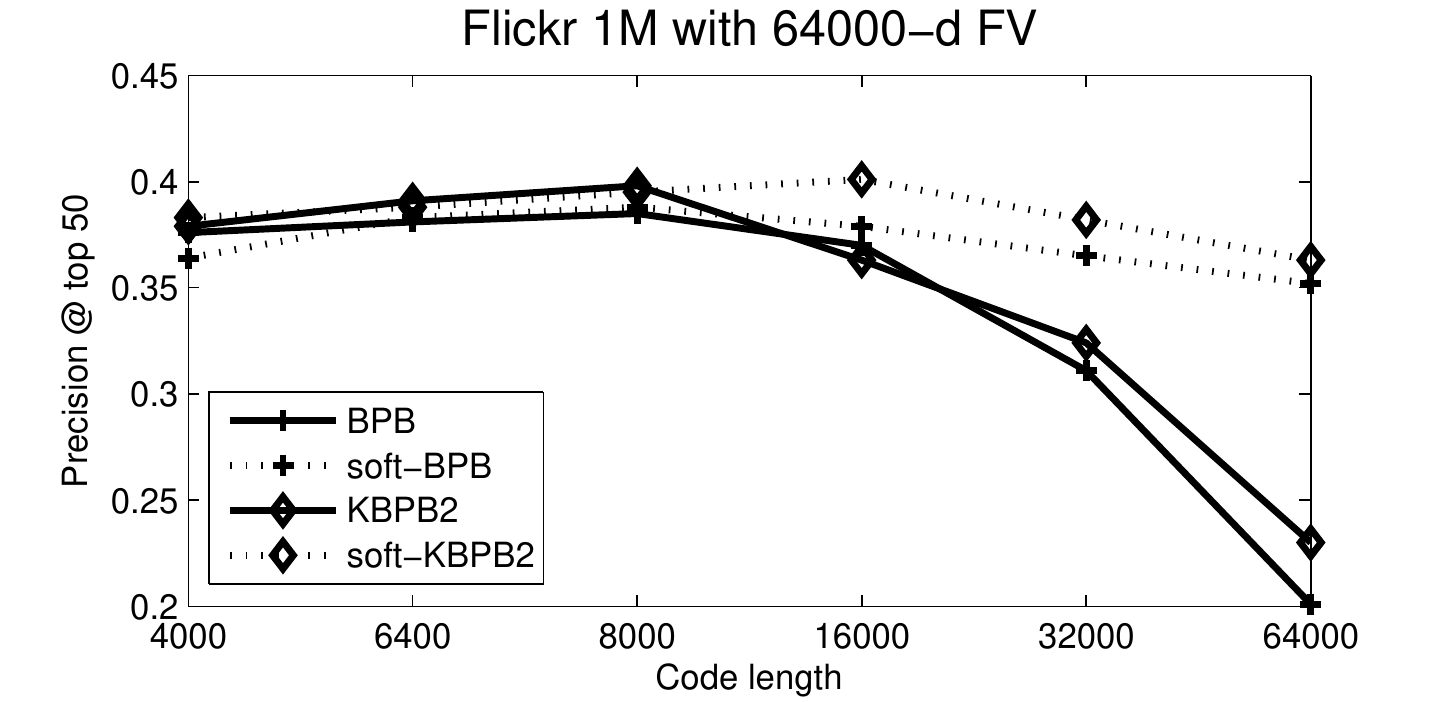}\\
  \caption{Comparison with soft-kmeans on BPB}\label{SOFT}
  \vspace{-2ex}
\end{figure}

\begin{figure*}
  \centering
  \begin{tabular}{cccc}
     \includegraphics[width=0.21\textwidth]{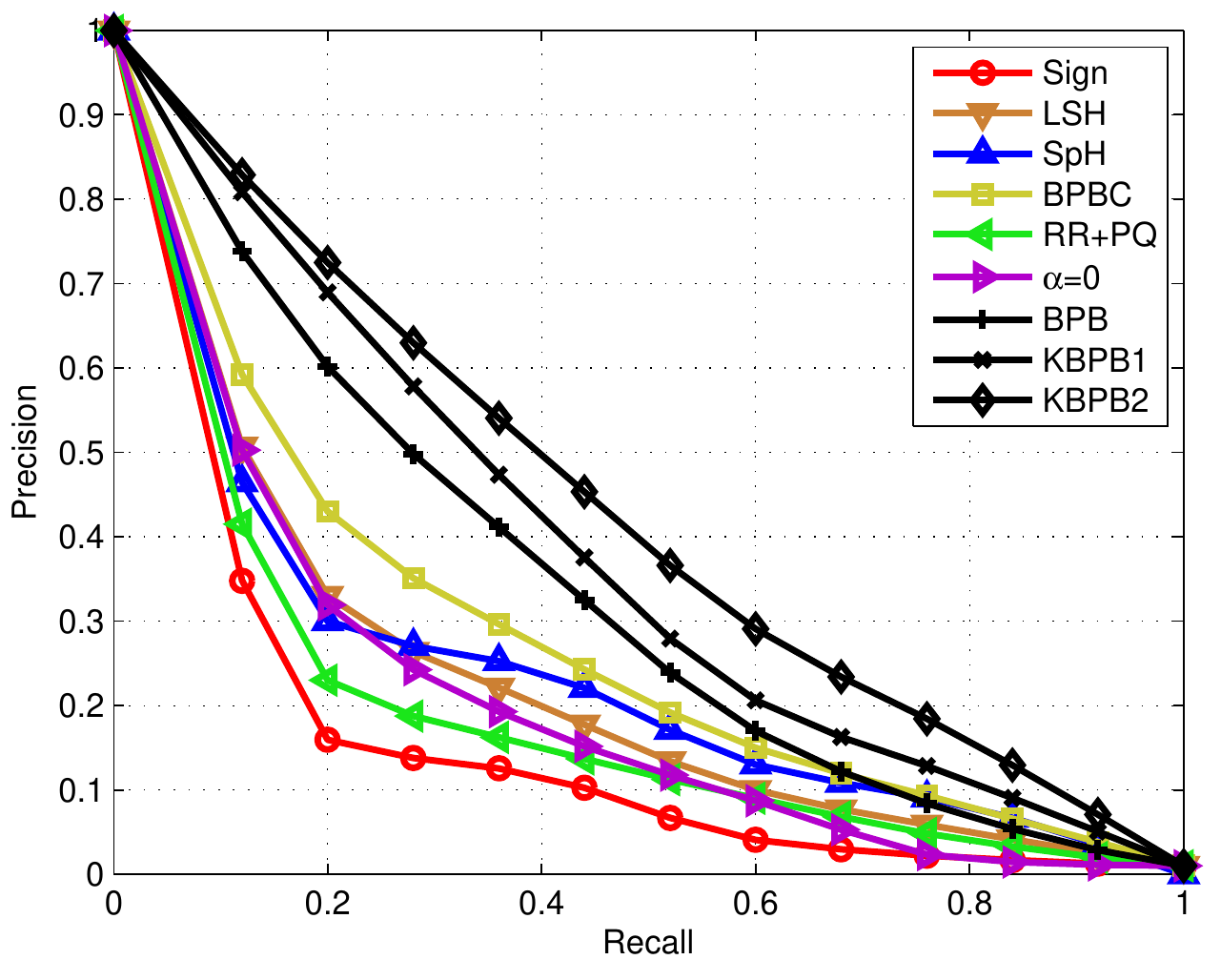} & \includegraphics[width=0.21\textwidth]{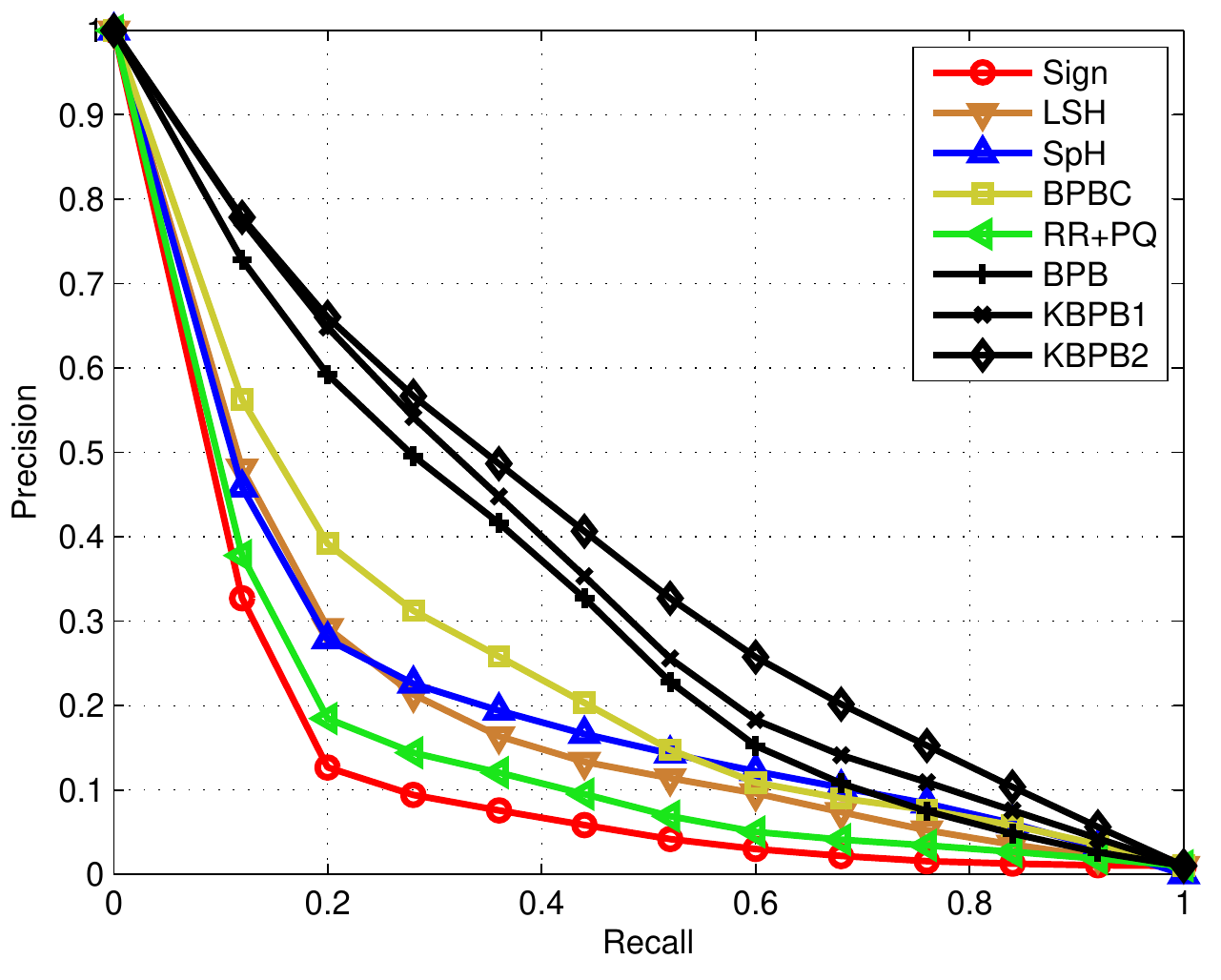} & \includegraphics[width=0.21\textwidth]{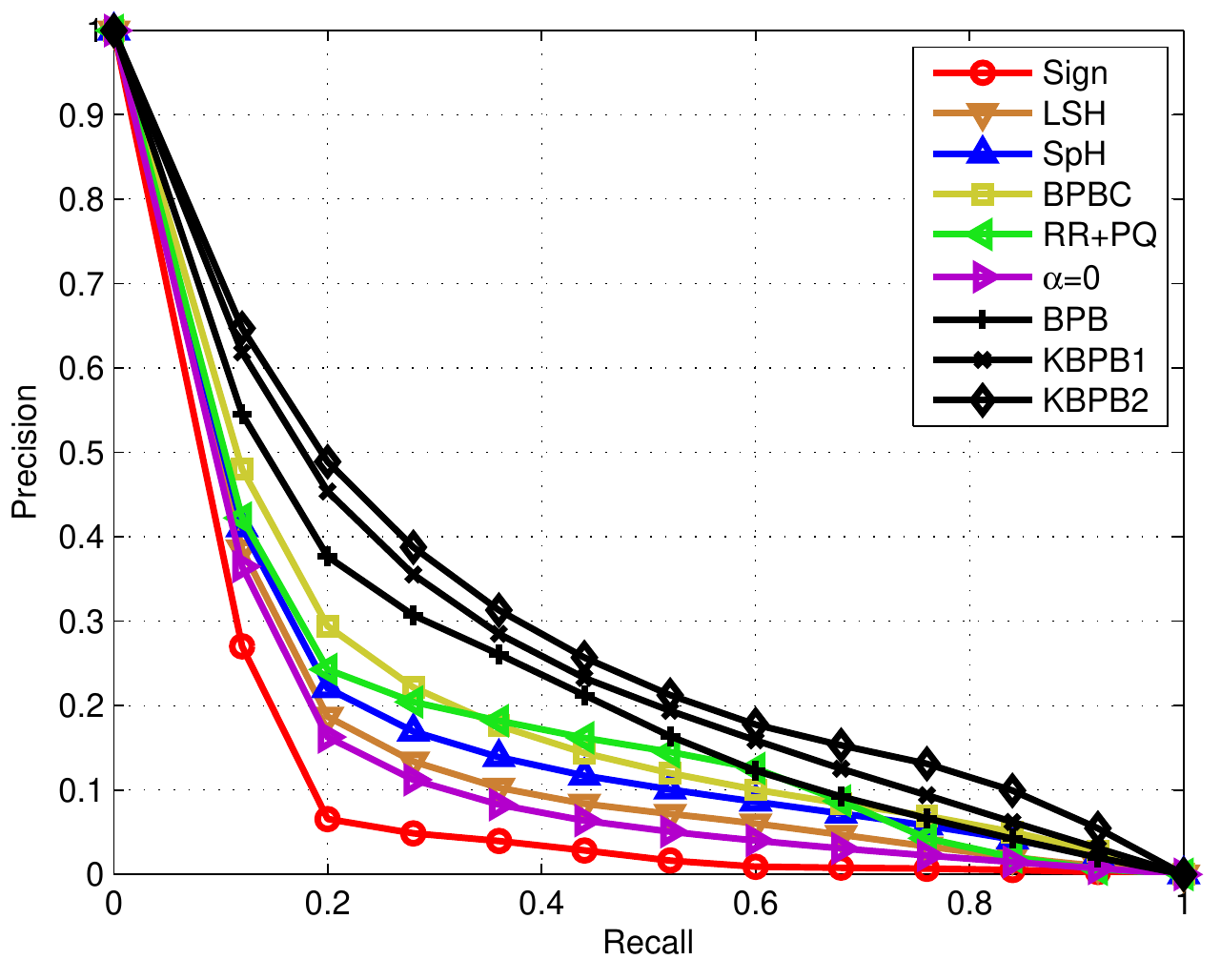} &
     \includegraphics[width=0.21\textwidth]{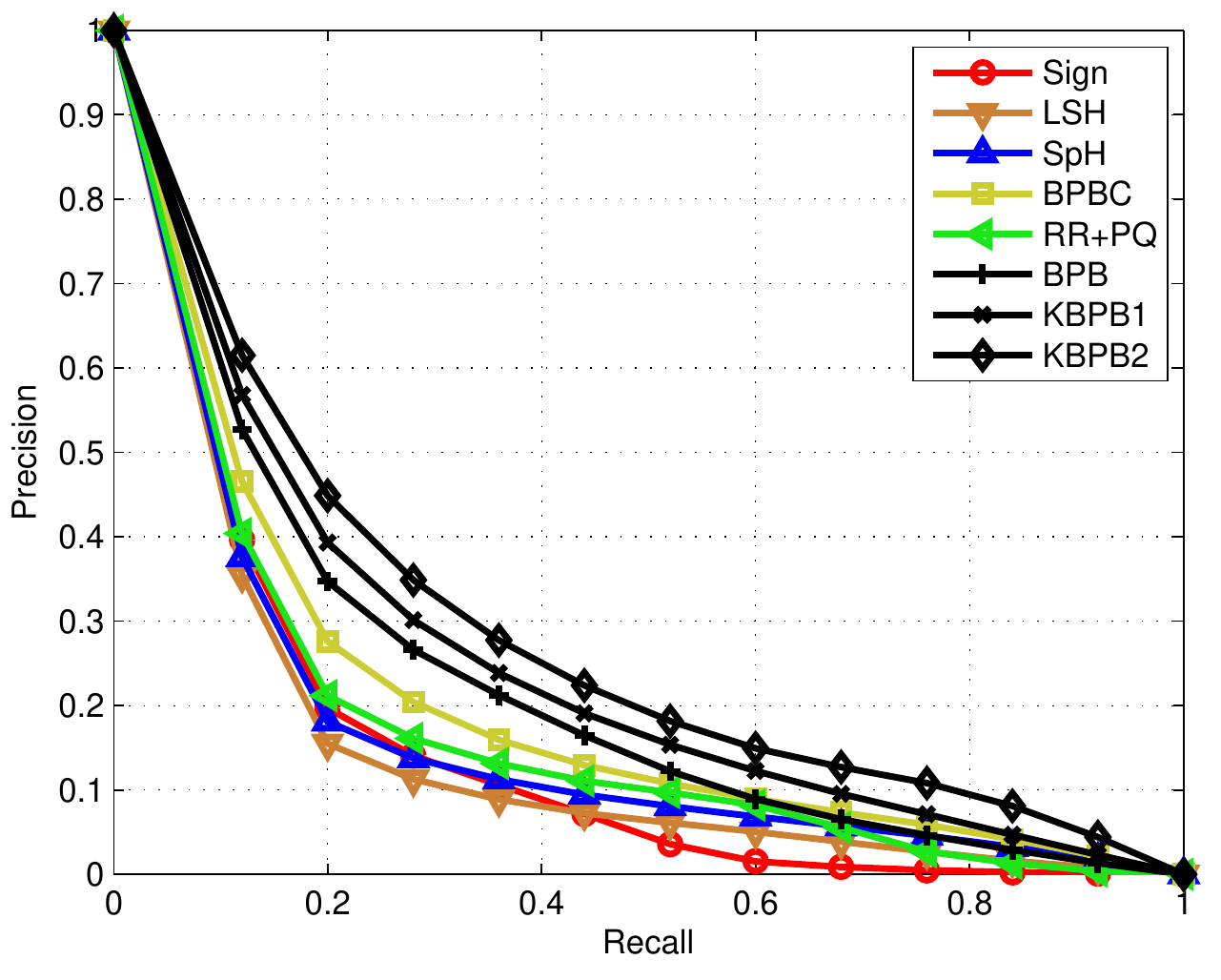}\\
    (a) Flickr 1M (FV) & (b) Flickr 1M (VLAD) & (c) ILSVR2010 (FV) & (d) ILSVR2010 (VLAD)  \\
  \end{tabular}
  \caption{Comparison of precision-recall curves on Flickr 1M and ILSVR2010 datasets with 8000 bits FV and 4000 bits VLAD.}
  \label{f4}
  \vspace{-2ex}
\end{figure*}

From Table~\ref{table:t1} and Table~\ref{table:t2}, our BPB algorithm consistently outperforms all the compared methods at every code length and leads to competitive accuracies with  CBE and original FV/VLAD. Moreover, KBPB can achieve better performance than BPB since the kernel method can theoretically and empirically solve the problem of linear inseparability of subspaces with relatively lower dimensions (average dimension of each subspace is $D/d$). Thus, KBPB gives significantly better performance when $d$ is large, i.e., on relatively long binary codes. The best performance on both datasets has been achieved by KBPB with the RBF kernel. Especially, when the code length decreases, the retrieval accuracies from all compared methods (expect SpH) dramatically drop, but the accuracies of our methods only slightly change showing the robustness of the proposed methods on \emph{medium-dimensional} binary coding.  Currently, we use hard-assignment K-means for our work. In Fig.~\ref{SOFT}, we have also evaluated the possibility to use soft-assignment clustering for our methods. The results illustrate that for the medium-dimensional codes (i.e., between 1000 bits and 10000 bits), soft-assignment clustering based BPB and RBF-KBPB can achieve competitive results with hard-assignment BPB and KBPB. However, in the extreme condition (i.e., code dim$\rightarrow$feature dim), the soft-clustering based methods can still produce the reasonable results without lossing much of accuracy, while the current hard-clustering methods fail, since the feature dimensions can be re-used during soft-clustering. Thus, from Fig.~\ref{SOFT}, Table~\ref{table:t1} and Table~\ref{table:t2}, we can observe our current version of projection bank can indeed achieve better performance for the  medium-dimensional codes compared with other methods. Besides, Fig.~\ref{f4} presents the precision-recall curves of all compared methods on both datasets with 8000 bits for FV and 4000 bits for VLAD, respectively. From all these figures, we can further discover that, for both datasets, BPB/KBPB outperform other high-dimensional compression methods with the medium-lengthed codes by comparing the retrieval precision and the Area Under the Curve (AUC).

\begin{table}
\centering
\newcommand{\tabincell}[2]{\begin{tabular}{@{}#1@{}}#2\end{tabular}}
\tiny
\caption{ Comparison of computational cost on different code lengths for 64000-d FV reduction on ILSVR2010 (Data stored in \emph{double} precision).}
\label{table:t3}
\begin{tabular}{|c|c|c|c|c|c|c|c|}
\hline
\textbf{\tabincell{c}{Code\\length}}&\textbf{Measurement}&RR & RR+PQ &BPBC & CBE&\textbf{BPB}&\textbf{KBPB}\\
\hline
3200&\textbf{Projections}& 1562.53 &1576.37 &0.34& 0.49&0.49&24.40\\
4000&\textbf{storage}&1953.19 &1970.49 &0.42 & 0.49 &0.49&30.51\\
6400&\textbf{(MB)}&3125.04 &3152.50 & 0.65&  0.49&0.49&48.83\\
8000& &3906.38&3940.38 & 0.80& 0.49 &0.49&61.17\\
\hline
 3200& \textbf{Training}& 2.21 & 1102.72& 848.21& 304.15& 314.27& 603.52\\
 4000& \textbf{time}& 3.30 & 1563.48& 1021.32& 340.24& 389.56& 718.33\\
 6400& \textbf{(s)}&  4.52 & 1918.52& 1336.10& 422.44& 472.33& 802.51\\
 8000&& 8.64 & 2204.30& 1559.08& 464.23& 505.16& 890.24\\
\hline
3200&\textbf{Coding}&246.71 & 599.47 & 5.46& 21.64 &0.24&2.41\\
4000&\textbf{time}&370.24 & 791.43& 13.45& 22.32  &0.23&6.64\\
6400&\textbf{(ms)}&534.10 & 1182.57 &30.57&  21.03 &0.24&19.10\\
8000&&721.51& 1603.10 &80.02 & 23.15 &0.24&55.91\\
\hline
\end{tabular}
\\ The training and coding time in this table are both 50-run averaged runtime.  Parallel computation is adopted to speed up our training phase.
\vspace{-8ex}
\end{table}

\begin{figure*}
  \centering
  \begin{tabular}{cccc}
     \includegraphics[width=0.21\textwidth]{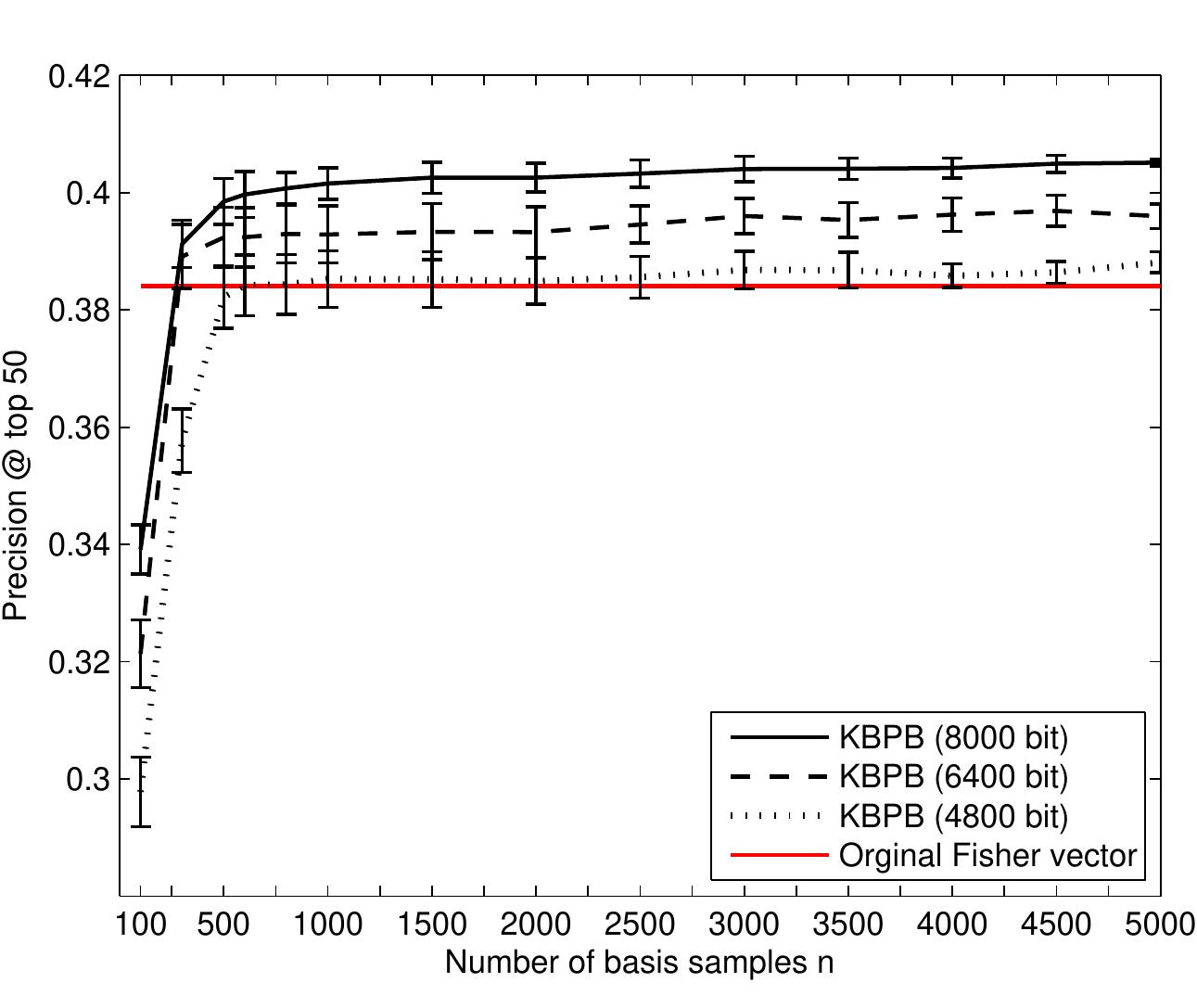} & \includegraphics[width=0.21\textwidth]{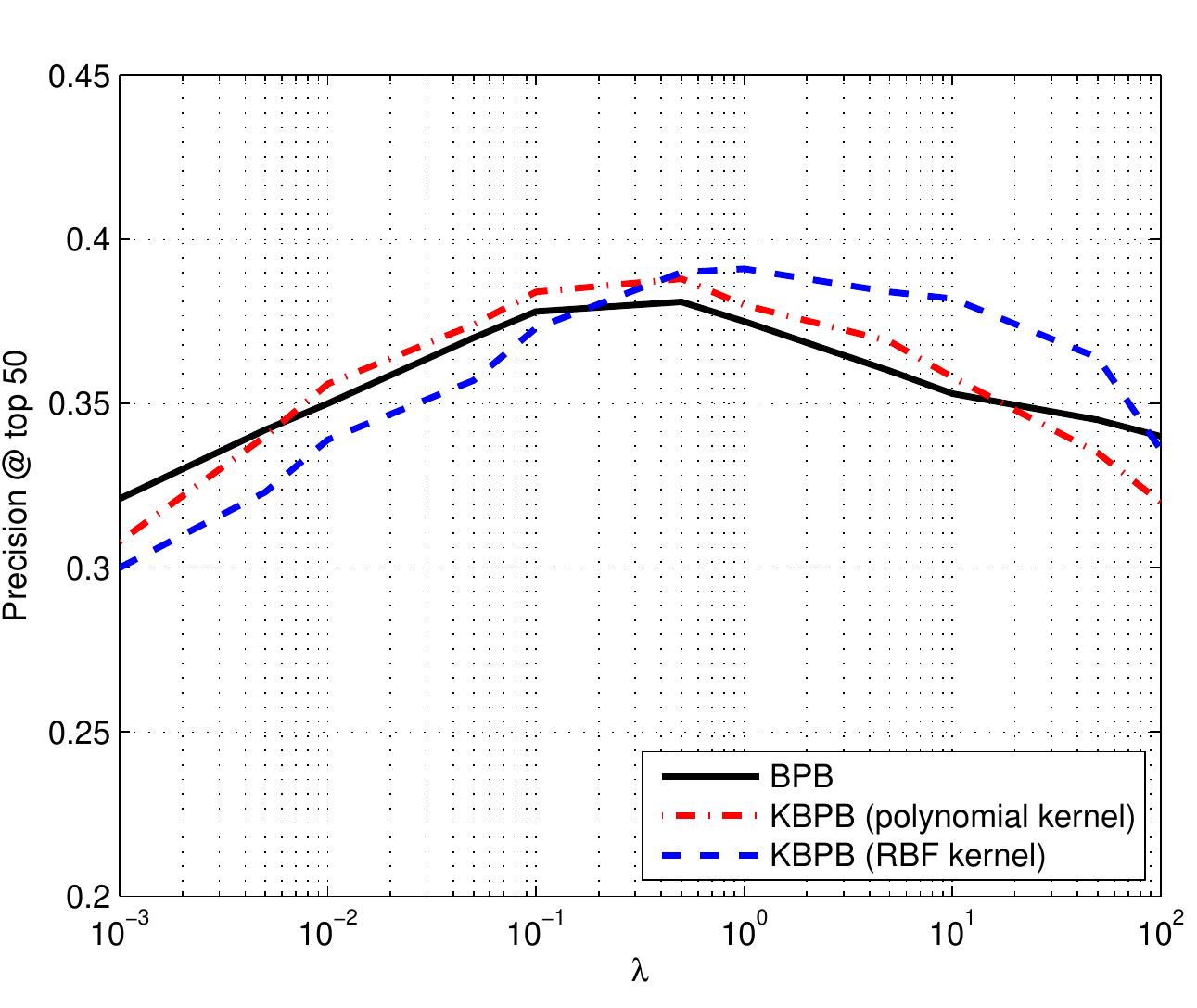} & \includegraphics[width=0.21\textwidth]{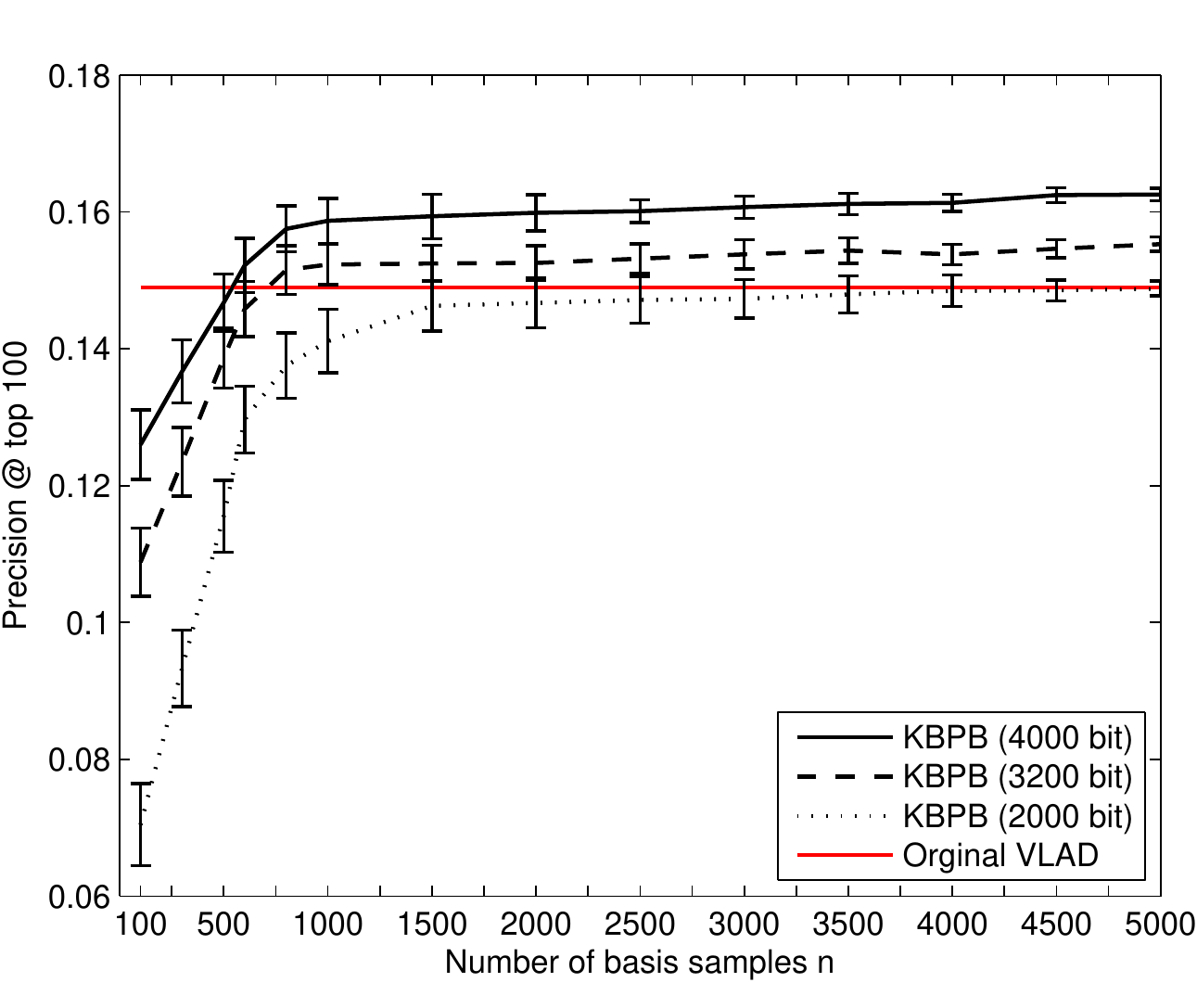} &
     \includegraphics[width=0.21\textwidth]{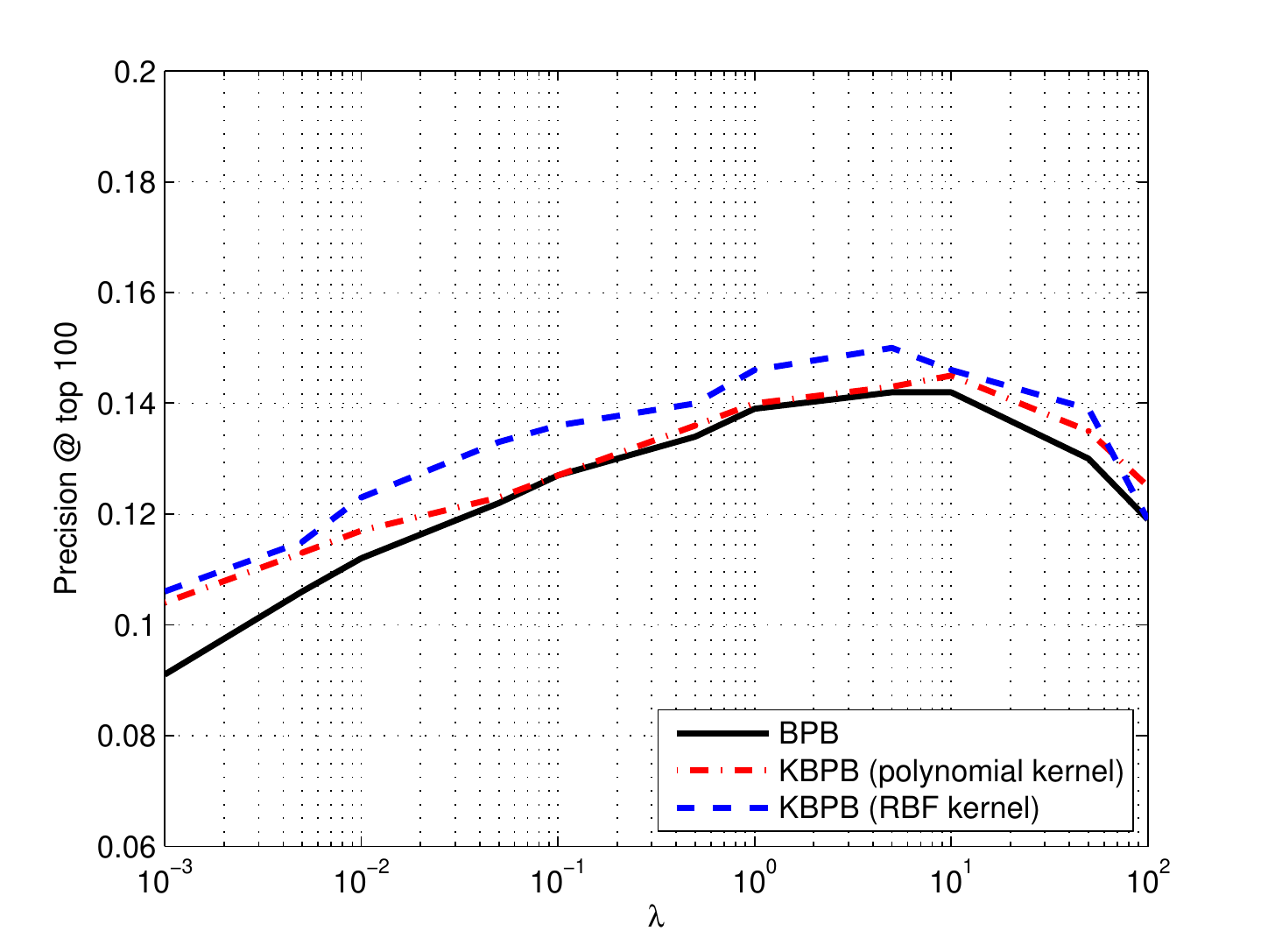}\\
    (a) Flickr 1M (FV) & (b) Flickr 1M (FV) & (c) ILSVR2010 (VLAD) & (d) ILSVR2010 (VLAD)  \\
  \end{tabular}
  \caption{(a) and (c) show the mean of 50 runs of retrieval accuracies of KBPB (with the RBF kernel) vs. parameter $n$ on Flickr 1M and ILSVR2010. (b) and (d) show the parameter sensitivity analysis of $\lambda$ on Flickr 1M and ILSVR2010 at 6400 bits and 3200 bits, respectively.}
  \label{f3}
    \vspace{-3ex}
\end{figure*}

\vspace{-3ex}
\paragraph{Complexity and parameter sensitivity analysis:}  Table \ref{table:t3} illustrates the comparison of the memory usage for projections, the training time and the coding time on ILSVR2010. RR+PQ costs the largest memory space and more time for coding since the full-matrix projection (i.e., RR) is involved. Compared with RR+PQ, BPBC and  CBE need much lower memory costs and time complexity for training and coding. Our BPB is slightly time-consuming than CBE in the training phase but the most efficient one for coding. Meanwhile, KBPB costs more memory space than BPBC and BPB but is still more efficient for coding than BPBC. In addition, Fig.~\ref{f3} reports the effect of performance by varying two essential parameters $n$ and $\lambda$. In terms of the number of basis samples $n$ used in \nobreak KBPB with the RBF kernel, when $n \geq 1000$, the retrieval accuracy curves become approximately stable on both datasets with FV and VLAD, respectively. It indicates that our KBPB can lead to relatively robust results with $n \geq 1000$.  As we can see, for balance parameter $\lambda$, our methods (both BPB and KBPB) can achieve the good performance when $\lambda \in (10^{-1}, 1)$ and $\lambda \in (1, 10)$ on Flick 1M and ILSVR2010, respectively.

\vspace{-0.5ex}
\subsection{Large-scale action recognition}
Finally, we evaluate our methods for action recognition on the UCF101 dataset \cite{soomro2012ucf101} which contains 13320 videos from 101 action categories. We strictly follow the 3-split train/test setting in \cite{soomro2012ucf101} and report the average accuracies as the overall results. The 426-dimensional default Dense Trajectory Features (DTF) \cite{wang2011action} are extracted from each video, and GMM and K-means are used to cluster them into 200 visual words for FV and VLAD respectively. Thus, the length of FV is $2\times200\times426=170400$ and the length of VLAD is $200\times426=85200$. For our methods, we fix $n=500$ and $\lambda=8$, which are both selected via cross-validation set, and other parameters are the same as the previous retrieval experiments. In this experiment, we apply the linear SVM\footnote{According to \cite{sanchez2011high,gong2013learning}, hashing kernel \cite{shi2009hash,weinberger2009feature} renders to an unbiased estimation of the dot-product in the original space. Thus, binary codes can also be directly fed into a linear SVM.} for action recognition. From the relevant results shown in Table~\ref{table:t5}, it can be observed that the recognition accuracies computed by all methods have generally smaller differences compared with the diversity of performance in retrieval tasks. The reason is that the supervised SVM training can compensate the discriminative power between different methods, whereas the unsupervised retrieval cannot. Our BPB and KBPB can not only achieve competitive results with original features, but also perform better than other compression methods on  medium-lengthed codes with FV and VLAD. Moreover, KBPB$^{2}$ consistently gives the best performance.

\begin{table}
\centering
\newcommand{\tabincell}[2]{\begin{tabular}{@{}#1@{}}#2\end{tabular}}
\setlength{\baselineskip}{7pt}
\fontsize{5pt}{\baselineskip}\selectfont
\caption{Comparison of action recognition performance (\%) on the UCF 101 dataset.}
\label{table:t5}
\begin{tabular}{|c||c|c|c|c|c|c|}


\cline{1-7}
\multirow{2}{*}{\textbf{Methods}}&\multicolumn{3}{c|}{\textbf{Fisher Vector} (170400-d)} &
\multicolumn{3}{c|}{\textbf{VLAD} (85200-d)}\\

\cline{2-7}
& \textbf{17040 bit} & \textbf{11360 bit} & \textbf{8520 bit} &\textbf{8520 bit} & \textbf{5680 bit} & \textbf{4260 bit}  \\
\hline
\hline
\rowcolor{LC}
Original &80.33 &80.33 &80.33 &77.95  &77.95&77.95\\
\hline
PCA &78.62 &78.31 &75.4 &77.03  &76.28 &74.1  \\
RR+PQ&77.25 &77.67 &75.50 &75.38  & 75.21& 74.03 \\
PKA &80.30 &78.88 &76.54 &77.21  &77.00 &76.4  \\
PQ&75.90 &74.84 &74.31 &72.85 &72.01  & 70.99 \\
\hline
sign &75.26& 75.26& 75.26& 74.41&  74.41& 74.41  \\
$\alpha=0$&76.78 &75.20 &74.56 &- &-  &-   \\
LSH &74.19 & 73.02 &71.88& 72.40  & 71.11& 70.4 \\
SpH & 71.36& 73.04 & 75.28 & 69.35  & 72.97 & 74.83  \\
BPBC & 77.21 & 76.40 & 75.89 & 75.91  & 74.73 & 73.22  \\
 CBE & 80.65 & 78.23 & 76.47 & 77.91  & 75.34 & 74.03\\
\hline
\hline
\textbf{BPB}  &80.02&79.26 &78.30 &77.53 &76.38  &75.52  \\
\textbf{KBPB}$^{1}$ &80.74& 80.35& 79.37& 78.28&  77.31& 76.54 \\
\rowcolor{Red}
\textbf{KBPB}$^{2}$ &82.18&81.55 &80.71&78.69&77.52  &76.90   \\
\hline
\end{tabular}
\vspace{-6ex}
\end{table}

\section{Conclusion and Future Work}
In this paper, we have presented a novel binarization approach called Binary Projection Bank (BPB) for high-dimensional data, which exploits a group of small projections via the max-margin constraint to optimally preserve the intrinsic data similarity. Different from the conventional linear or bilinear projections, the proposed method can effectively map very high-dimensional representations to \emph{medium-dimensional} binary codes with a low memory requirement and a more efficient coding procedure. BPB and the kernelized version KBPB have achieved better results compared with state-of-the-art methods for image retrieval and action recognition applications. In the future, we will focus more on using soft-assignment clustering based projection bank methods.

{\small
\bibliographystyle{ieee}
\bibliography{iccv2015}
}

\end{document}